\documentclass[conference,a4paper]{IEEEtran}

\ifCLASSINFOpdf
  \usepackage[pdftex]{graphicx}
  \graphicspath{{../figures/}}
  \DeclareGraphicsExtensions{.pdf,.jpeg,.png}
\fi

\newcommand\blfootnote[1]{%
  \begingroup
  \renewcommand\thefootnote{}\footnote{#1}%
  \addtocounter{footnote}{-1}%
  \endgroup
}

\usepackage{graphicx}
\usepackage[export]{adjustbox}
\usepackage{amsmath,amssymb} 
\usepackage[colorlinks]{hyperref}
\usepackage{cleveref}
\usepackage{array}
\usepackage{caption}
\usepackage{multirow}

\usepackage{color}
\usepackage{array}
\usepackage{authblk}

\begin{document}

\title{Deep Latent Space Learning for Cross-modal Mapping of Audio and Visual Signals}

\author[1]{Shah Nawaz\textsuperscript{*}}
\author[2]{Muhammad Kamran Janjua\textsuperscript{*}}
\author[1]{Ignazio Gallo}
\author[3]{Arif Mahmood}
\author[1]{Alessandro Calefati}
\affil[1]{Department of Theoretical and Applied Science, University of Insubria, Varese, Italy}
\affil[2]{SEECS, National University of Sciences and Technology, Islamabad, Pakistan}
\affil[3]{Department of Computer Science, Information Technology University, Lahore, Pakistan}
\affil[ ]{\tt\small {\{snawaz,ignazio.gallo,a.calefati\}@uninsubria.it}, {mjanjua.bscs16seecs@seecs.edu.pk}}

\affil[ ]{\tt\small {arif.mahmood@itu.edu.pk}}


\maketitle
\begin{abstract}
We propose a novel deep training algorithm  for joint representation of audio and visual information\blfootnote{*Equal contribution}
which consists of a single stream network (SSNet) coupled with a novel loss function to learn a shared deep latent space representation of multimodal information. The proposed framework characterizes the shared latent space by leveraging the class centers which helps to eliminate the need of pairwise or triplet supervision.
We quantitatively and qualitatively evaluate the proposed approach on VoxCeleb, a benchmarks audio-visual dataset on multitude of tasks including cross-modal verification, cross-modal matching and cross-modal retrieval. State-of-the-art performance is achieved on cross-modal verification and matching while comparable results are observed on the remaining applications. Our experiments demonstrate the effectiveness of the technique for cross-modal biometric applications.
\end{abstract}


\section{Introduction}
\textit{Why do people watching a large screen in a movie theater hear an actor's voice coming from his face, even though the audio speakers are on the side of the hall?} The illusion of third voice due to association of visual data with the auditory information is known as the McGurk effect~\cite{mcgurk1976hearing}. It is not surprising that auditory experience is influenced by the visual input. This is the case with the ventriloquist who exploit the visual capturing ability of the audience to sell the idea that their puppets speak. Since the audience sees the puppet moving its lips, the location of sound also changes simultaneously and the audience experiences the McGurk effect.

It is a well studied fact that humans end up associating voices and faces of people~\cite{kamachi2003putting,belin2004thinking} due to the fact that neuro-cognitive pathways for voices and faces share the same structure~\cite{ellis1989neuro}. This behavioral tendency is best exploited in a Hollywood film where a person can be identified just by hearing his/her voice. In fact a movie named `Taken' is based on the identification of a person using only the auditory input. Recognition of speakers from their voices requires information from cross domains to be mapped onto a shared latent space~\cite{nagrani2018seeing}. 
Once the joint representation is obtained, a family of algorithms can be introduced, ranging from matching, verifying, authenticating, retrieving and even searching.  
However, speaker recognition under unconstrained conditions is an exceedingly difficult task, since real-world scenarios are not limited to noise-free environments. Background noises play an important role and have become an inevitable part of our everyday life. Without characterizing the background noise and variation, holistic understanding and robust learning is not possible~\cite{nagrani2017voxceleb}. 
 
The task of cross-modal mapping is supported by the hypothesis presented by studies above that it may be possible to find association between voices and faces as well. With this in perspective, the current paper focuses on the task of obtaining a joint representation of auditory and visual input employing a single stream network for both modalities. The problem setup is that we have a corpus of auditory and visual information (voices and faces) and we perform following tasks on the dataset using a single network: \textit{matching, retrieval and verification}. 

One of the major hindrance in performing such experiments is the unavailability of large-scale corpus consisting of information from both domains (images and audio). However, recently VoxCeleb dataset~\cite{nagrani2017voxceleb} has been introduced which comprises of a collection of video and audio recordings of a large number of celebrities. Previous works in literature~\cite{kim2018learning,nagrani2018seeing,nagrani2018learnable} have modeled the problem of cross modal matching by employing separate networks for multiple modalities in either triplet network fashion or subnetwork. Separate networks in triplet fashion may help with modularity given few modalities (two in this case) at input, but it is important to take into account the possibility of multiple input streams (text, image, voice, etc). In a triplet fashion, each modality acquires $\mathcal{O} (n^3)$ space complexity, where $n$ is sample size. In VoxCeleb dataset, there are $1,251$ identities which becomes $2,502$ if we consider single image and single voice for each identity; the space complexity grows exponentially if we increase number of instances of each of each identity. 

Nagrani et. al~\cite{nagrani2018seeing} performed experiments in static and dynamic settings. A five stream dynamic-fusion architecture\footnote{The five stream dynamic-fusion architecture consists of two face sub-networks, one voice network along with two extra streams as dynamic feature subnetworks} requires five subnetworks to account for this fusion. 
We achieve the five stream dynamic-fusion results with a single network trained in end-to-end fashion. Our network performs under no restrictions in terms of triplet selection. We perform series of experiments inspired from~\cite{nagrani2018seeing,wen2018disjoint}. Furthermore, we perform two additional experiments to establish the robustness of our methodology. Our main contributions are listed below.
\begin{itemize}
\itemsep0em 
\item [--] We introduce a single, end-to-end trainable network for performing auditory and visual information matching, verification and retrieval. 
\item [--] We propose a novel training procedure which can be coupled with deep neural networks to map multiple modalities to shared latent space without pairwise or triplet information. 
\item [--] We perform series of cross modal matching, verification and retrieval experiments considering multiple demographic, age, and gender factors in the wild.
\end{itemize}

The rest of the paper is structured as follows: we explore the related literature in Section~\ref{sec:related-work}; details of the proposed approach are discussed in Section~\ref{sec:proposed-approach}, followed by experiments and evaluations in Section~\ref{sec:experiments} and Section~\ref{sec:evaluation} respectively. Section~\ref{sec:discussion} presents discussion, followed by ablation study in Section~\ref{sec:abl-study}. Finally, conclusions are in Section~\ref{sec:conclusion}.

\section{Related Work}
\label{sec:related-work}
In this section we study previous works under multiple subsections. We lay the foundation of the problem in cognitive neuroscience studies followed by the introductory work done by the vision community. 
\subsection{Cognitive Neuroscience Studies}
The characteristic display of auditory information in emotional analysis dates back to the times of Darwin. In his book \textit{The expression of emotions in man and animals}, Darwin recognized emotions as non-private experiences and described their characteristic displays. Speech perception studies in psychology provide experimental support to Darwin's idea of an existence of link between auditory and visual information~\cite{driver1996enhancement,mcgurk1976hearing,yehia1998quantitative}. Studies like~\cite{kamachi2003putting,belin2004thinking} provide evidence that equivalent information about identity is available cross-modally from both the auditory and visual information domains. The McGurk effect illustrates the fundamental idea \textit{what is seen can affect what is heard}. It also explains how ventriloquists enhance spatial attention to sound by employing the cross modal information from the visual cues. More generally, the study in~\cite{vatikiotis1998moving} explains that having the speaker's face visible improves the perception of speech in noise. It is important to note that human perception studies have performed static matching of voices and faces, but argue that the results lie at chance level~\cite{lachs2004specification,kamachi2003putting}. We see that these studies have been conducted previously under the cognitive psychological perspective in detail and have recently been introduced to the vision community.
\subsection{Audio and Visual Recognition}
Audio and visual recognition are important tasks and main stream vision techniques have been focused towards solving the problem. 
In recent years, this problem of face and audio recognition has seen extensive progress~\cite{calefati2018git,parkhi2015deep,taigman2014deepface,guo2016ms,snyder2017deep,dehak2011front}.
It is important to note that these approaches allow for effective representation of unimodal information. However, the alignment of information learned across modalities is not accounted for in these approaches. In the current work we present a single stream architecture trained with the distance learning paradigm to obtain effective representations of audio and visual signals in the shared latent space.
\subsection{Joint Latent Space Representation}
To effectively capture cross modal embeddings, information across all the modalities have to be learned and mapped onto a joint latent space. Cross modal learning employing visual and textual data has seen significant progress~\cite{frome2013devise,vendrov2015order,zitnick2013learning,antol2015vqa}. However, not much work has been directed towards exploring joint representation of audio and visual data. Few works focus on the tasks of audio-visual matching for scene understanding~\cite{arandjelovic2017look,aytar2018cross,aytar2016soundnet,tian2018audio,torfi20173d}. Recently, a dataset tailored towards audio-visual biometrics was introduced~\cite{nagrani2017voxceleb,Chung18b} to aid the learning of audio and visual information and thus obtaining a joint representation. Many works have been focused towards speaker recognition and matching from audio and visual signals~\cite{nagrani2018seeing,nagrani2018learnable}. Although these works effectively capture cross modal embeddings, however they require either separate networks for each modality and/or require pair selection during training to effectively penalize the negative pairs. Considering the data explosion, pair selection in the wild becomes impractical. Although work in~\cite{wen2018disjoint} does not require pair selection, it requires sequential input to two separate networks with covariate knowledge which is obtained via a multi-task classification network conditioned under covariate supervision to obtain a joint representation of the audio and visual signals. In the current work we remove the overhead of pair selection or requirement of knowledge of any covariate or separate networks for each modalities. Instead we propose a single stream network to capture joint representation in the shared latent space employing a distance learning approach.
\section{Deep Latent Space Learning Approach}
\label{sec:proposed-approach}
\begin{figure*}[!t]
 \center
 \includegraphics[width=0.95\textwidth]{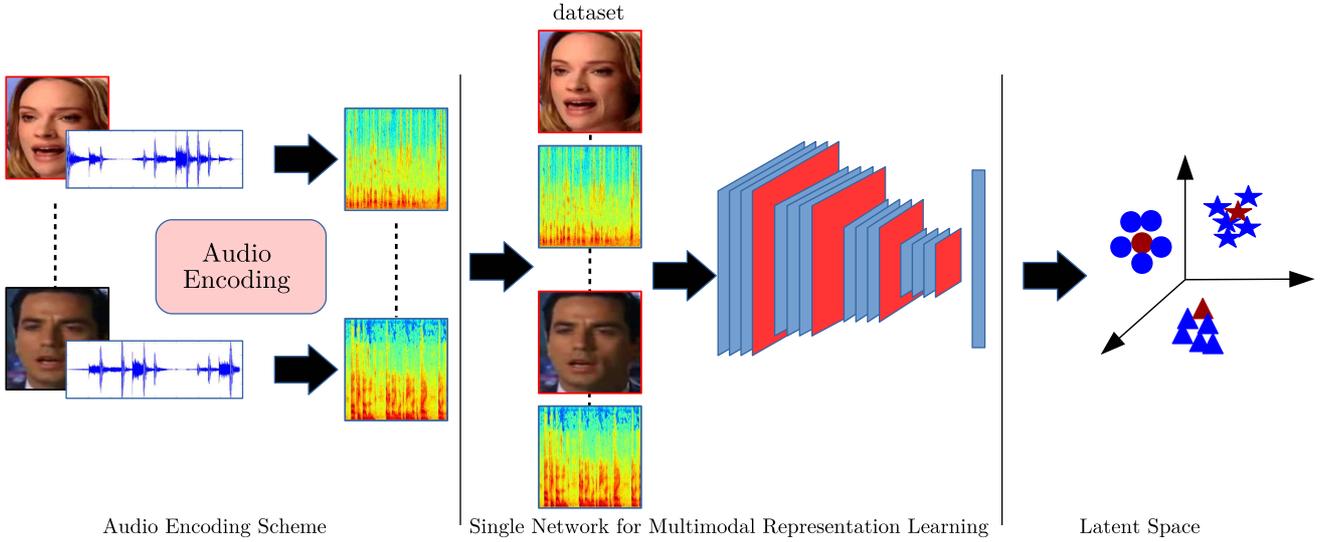}
  \caption{The proposed cross-modal framework based on single stream network and a novel loss function to embed both audio and visual signals in the latent space. The single stream network extracts the representation from both modalities while the loss function learns to bridge the gap between them.}
  \label{fig:arch}
\end{figure*}

One of the core ideas of this paper is to bridge the gap between the images and encoded voice signals i.e. spectrograms.  Our proposed approach eliminates the need for multiple networks for either modality, since similar results can be achieved with a single network. Fig.~\ref{fig:arch} visually explains the framework of the proposed approach.
%
%
Consider $\{A_j, I_j,y_j\}_{j=1}^N$ be $N$ training data points where $A_j$ and $I_j$ represent audio and visual signals belonging to a class $j$ and $y_j$ represents the labels. The objective of latent space learning is based on minimization of distance $D_\theta(A_j, I_j)$ between features and center of class $j$. 
This can be achieved with the help of a single stream convolutional neural network which is trained in an end-to-end fashion.
The details of the proposed approach are presented in the following  subsections: 

\subsection{Visual Signals}
The input to single stream network consists of three channel (RGB) facial image cropped to represent only the facial region. The input image size is fixed to $256\times256$ pixels. Note that under the proposed framework, single stream network is capable of handling both dynamic and static input regardless of any conditioning. 
\subsection{Audio Signals}
In addition to visual input, audio signals are also fed to the network. The encoded audio signals are short term magnitude spectrograms generated directly from raw audio of length three seconds. The audio stream is extracted, converted to single channel at 16kHz sampling rate with sampling frequency in accordance to the frame rate. The methodology is explained in~\cite{nagrani2017voxceleb,nagrani2018seeing}, however, we do not perform any normalization as a part of pre or post processing.

\subsection{The Single Stream Network} 
Our proposed network is generic and an appropriate deep neural network can be employed. In our implementation, we use  InceptionResNet-V1 as a single stream network for joint embedding of audio and visual signals (Fig.~\ref{fig:arch}). The network is trained using both face images and spectrograms. Suppose there are $n_s$ audio spectrograms associated with $n_i$ face images in a class $c$ representing an identity. Each image and the spectrogram is input to the network and $n_s+n_i$ feature vectors $f_c$ are obtained at the output of the network. During training, geometric center of $n_s+n_i$ feature vectors is computed and the objective function consisting of the distance of each feature vector from the center, is minimized. 
\begin{equation}
\label{eq:centerloss}
d(f_c) = \sum_{i=1}^{n_s+n_i}\parallel f^i - \frac{1}{n_s+n_i}\sum_{j=1}^{n_s+n_i}{f^j}\parallel_{2}^2
\end{equation}
Thus during the training phase, face images and the spectrograms are treated in similar fashion and a single stream network can effectively bridge the gap between image and audio  eliminating the need of multiple networks for each modality.
In our implementation, instead of using the traditional loss functions, we extend center loss for cross-modal distance learning jointly trained with softmax loss~\cite{wen2016discriminative}. This loss function simultaneously learns centers for all classes including face images and spectrograms in a mini-batch and minimzes the distances between each center and the associated images and spectrograms. It thus imposes neighborhood preserving constraint within each modality as well as across modalities. 
If there are $n$ classes in a mini batch $m$, the loss function is given by 
\begin{equation}
\label{eq:centerloss-softmax}
\mathcal{L}(\text{mini batch}) = -\sum_{i=1}^{m} \log \frac{e^{W_{y_{i}}^{T} f^i+b_{y_{i}}}}{\sum_{j=1}^{n} e^{W_{j}^{T} f^i+b_{j}}} + \frac{\lambda}{2}\sum_{i=1}^m d(f_c)
\end{equation}

In Eq.~\ref{eq:centerloss-softmax}, $f^i\in \mathbb{R}^{d}$ denotes the \textit{i}th deep feature, belonging to the $y_j$th class and
$d$ is the feature dimension.
$W_{j}\in \mathbb{R}^{d}$ denotes the \textit{j}th column of the weights
$W\in \mathbb{R}^{d \times n}$ is the last fully connected layer and \textbf{b} $\in \mathbb{R}^{n}$ is the bias term.
A scalar $\lambda$ is used for balancing the two loss functions. 
The conventional softmax loss can be considered as a special case of this joint supervision, if $\lambda$ is set to $0$~\cite{wen2016discriminative}.

This loss function minimizes the variation between face image and spectrogram within a class and effectively preserves the neighborhood structure. In this way, face image and spectrogram which do not belong to the same identity do not occur in the same neighborhood. Implementation details are explained in the next section.

\section{Experiments}
\label{sec:experiments}
We perform a series of experiments on various tasks consisting of \textit{cross-modal verification}, \textit{cross-modal matching} and \textit{cross-modal retrieval} to evaluate the embeddings learned by the single stream network under the proposed framework.
The experimental setup and dataset details are explained below.

\subsection{Experimental Setup} 
We perform three different experiments which are as below.
\subsubsection{\textit{Cross-Modal Verification}}
The first task is to perform \textit{cross-modal verification} where the goal is to verify if an audio segment and a face image belong to the same identity. Two inputs are considered i.e. face and voice and verification between the two depends upon a threshold on the similarity value. The threshold can be adjusted in accordance to wrong rejections of true match and/or wrong acceptance of false match. We report results on standard verification metrics i.e. ROC curve (AUC) and equal error rate (EER).
\subsubsection{\textit{Cross-Modal Matching}}
The second task consists of \textit{cross-modal matching} where the goal is to match the input modality (probe) to the varying gallery size $n_c$  which consists of the other modality. We increase $n_c$ to determine how the results change. For example, in $1:2$ matching task, we are given a modality at input, e.g. face, and the gallery consists of two inputs from other modality, e.g. audio. One of them contains a true match and other serves as an imposter input. We employ matching metric i.e. accuracy to report results. We perform this task in five settings where in each setting the $n_c$ is increased as $2,4,6,8,10$. 
\subsubsection{\textit{Cross-Modal Retrieval}}
Lastly, we evaluate the learned embedding on \textit{cross-modal retrieval}. Given a single modality input, the task is to retrieve all the semantic matches of the opposite modality. We perform this task for both Face $\rightarrow$ Voice and Voice $\rightarrow$ Face formulation. We report results in terms of $R@K$ which evaluates accuracy in terms of the first $K$ retrieved results against a query. 

\subsection{Dataset} 
Recently, Nagrani et. al~\cite{nagrani2017voxceleb} introduced a large-scale dataset of audio-visual human speech videos extracted `in the wild' from YouTube.
Nagrani et. al~\cite{nagrani2018learnable} created two train/test splits out of this dataset to perform various cross-modal tasks.
The first split consists of disjoint videos from the same set of speakers while the second split contains disjoint identities.
We train the model using these two training sets, allowing us to evaluate on both test sets, the first one for \textit{seen-heard} identities, and the second for \textit{unseen-unheard} identities.
Note that we followed the same train, validation and test split configurations as used in~\cite{nagrani2018learnable} for fair comparisons.


\subsection{Implementation Details}
We learn single stream network with standard hyper-parameters setting. The size of the input images and spectrograms is specified to $256\times256$ and output feature vector is $128-d$ extracted from the last fully connected layer of single stream network. For optimization we employ Adam optimizer~\cite{kingma2014adam} because of its ability to adjust the learning rate during training. We use Adam's initial learning rate of $0.05$ and employ weight decay strategy with decaying by a factor of $5\mathrm{e}{-5}$. The network is trained for $100$ epochs. The mini-batch size was fixed to randomly selected $45$ images and spectrograms. Training with mini-batch speeds up the process and helps with generalization.

\section{Evaluation}
\label{sec:evaluation}
\subsection{Cross-modal Verification}
In this section we report results of the single stream network on \textit{cross-modal verification} task, the aim of which is to determine whether an audio segment and a face image are from the same identity or not.  
Recently~\cite{nagrani2018learnable} used VoxCeleb dataset to benchmark this task under two evaluation protocols, one for \textit{seen-heard} identities and the other for \textit{unseen-unheard} identities.
We evaluate on the same test pairs\footnote{\url{http://www.robots.ox.ac.uk/~vgg/research/LearnablePins}}
provided in~\cite{nagrani2018learnable} for each evaluation.
More specifically, $30,496$ pairs from unseen-unheard identities and $18,020$ pairs from seen-heard identities are selected.
The results for cross-modal verification are reported in Table~\ref{tab:results-verification}.
We use area under the ROC curve (AUC) and equal error rate (EER) metrics for verification.
As can be seen from the table, our model trained from scratch outperformed the state-of-the-art work on seen-heard protocol and unseen-unheard protocol. 

Furthermore, we examine the effect of Gender (G), Nationality (N) and Age (A) separately, which influence both face and voice verification. It is important to note that~\cite{nagrani2018learnable} employed pre-trained network, whereas we trained the model from scratch. Our network outperformed on G, N, A and the combination (GNA) in \textit{seen-heard} formulation regardless of pre-trained network as a backbone, see Table~\ref{tab:results-demographic}. However, our network shows comparable results on \textit{unseen-unheard} formulation for N,A and GNA, whereas it outperformed on random and G regardless of pre-trained network, see in Table~\ref{tab:results-demographic}. 

\begin{table}[tb]
\caption{Cross-modal verification results on seen-heard and unseen-unheard configurations with model trained from Scratch.}
\begin{center}
\label{tab:results-verification}
\begin{tabular}{|lcc|}
\hline
& AUC \% & EER \% \\
 \hline
& \multicolumn{2}{c|}{Seen-Heard}\\
 \hline
Learnable Pins~\cite{nagrani2018learnable}  & 73.8  & 34.1   \\
Proposed SSNet   & \textbf{91.1}  & \textbf{17.2}    \\
 \hline
 & \multicolumn{2}{c|}{Un-seen-Un-heard}\\
  \hline
Learnable Pins~\cite{nagrani2018learnable} & 63.5  & 39.2                        \\
Proposed SSNet   &\textbf{78.8}  & \textbf{29.5}       \\

\hline
\end{tabular}
\end{center}
\end{table}

\begin{table*}[t]
\caption{Analysis of cross-modal biometrics under varying demographics for seen-heard and unseen-unheard identities. Note that SSNet has produced best results when trained from scratch.}
\centering
\begin{tabular}{|llccccc|}
\hline
Demographic Criteria & Configuration &Random & G & N & A & GNA \\
\hline
 & \multicolumn{6}{c|}{Seen-Heard (AUC \%)}\\
\hline
Learnable Pins~\cite{nagrani2018learnable} & Scratch& 73.8             & -             & -             &  -            & -   \\
Learnable Pins~\cite{nagrani2018learnable}&Pre-train & 87.0            & 74.2          & 85.9          & 86.6          & 74.0 \\
Proposed SSNet &Scratch & \textbf{91.2}   & \textbf{82.5} & \textbf{89.9} & \textbf{90.7} & \textbf{81.8}  \\
\hline
 & \multicolumn{6}{c|}{Unseen-Unheard (AUC \%)}\\
 \hline
Learnable Pins~\cite{nagrani2018learnable} &Scratch & 63.5          & -            & - &  -  & -   \\
Learnable Pins~\cite{nagrani2018learnable} & Pre-train& 78.5          & 61.1            & 77.2 &  \textbf{74.9}  & \textbf{58.8}   \\
Proposed SSNet & Scratch& \textbf{78.8} & \textbf{62.4}   & 53.1  &   73.5          & 51.4    \\
\hline
\end{tabular}
\label{tab:results-demographic}
\end{table*}

\subsection{Cross-modal Matching}
\begin{figure}[!t]
 \center
 \includegraphics[width=0.5\textwidth]{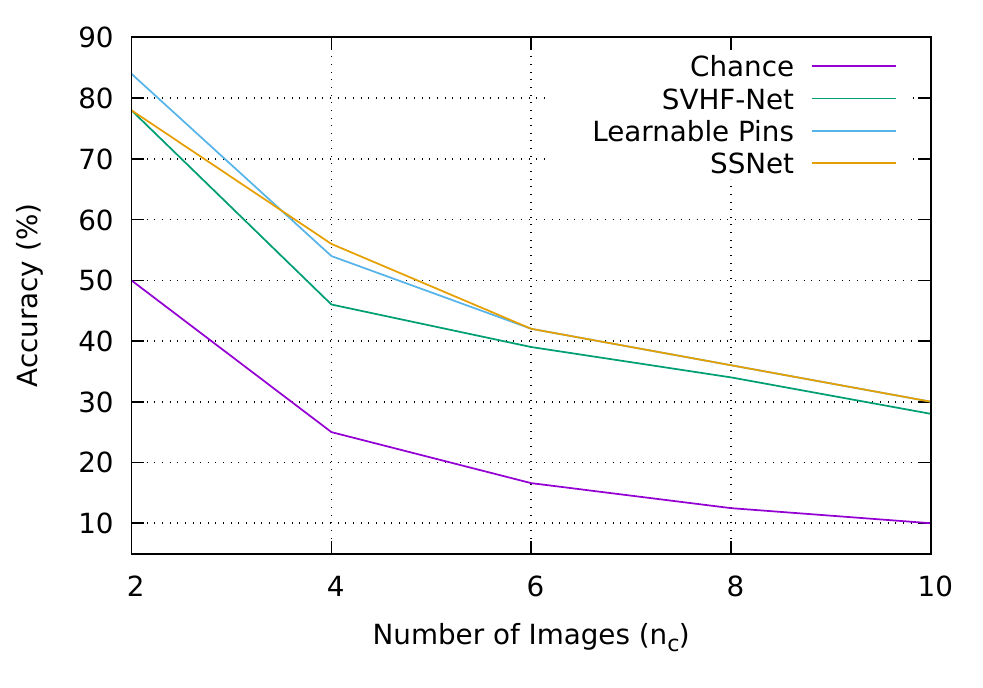}
  \caption{Qualitative results for Face $\rightarrow$ Voice matching task.}
  \label{fig:verification-fig}
\end{figure}

In this section we perform the \textit{cross-modal matching} task employing single stream network. We perform the $1:n_c$ where $n_c=2,4,6,10$ matching tasks to evaluate the performance of our approach. Unlike others~\cite{nagrani2018seeing,nagrani2018learnable}, we do not require positive or negative pair selection since under the proposed framework, the network learns in a self-supervised manner. Table~\ref{tab:nway} reports results on said task along with comparison with recent approaches on the same task. 
In Table~\ref{tab:nway}, the probe is voice while the matching gallery consists of faces. For instance consider the case where the input is voice and is $1:2$ matching task, we find that out the entry in gallery which best matches the input. It is important to note that for $n_c > 2$ tasks, the work in~\cite{nagrani2018seeing} trains separate network for each $n_c$. However, the major advantage of training under the proposed framework is that it is not restricted to a particular value of $n_c$. The single stream network can effectively handle any value of $n_c$ without increasing sub-network size. We observe that increasing $n_c$ decreases performance in a linear fashion due to increase in the challenge. Fig.~\ref{fig:verification-fig} shows the qualitative analysis of forced N-way matching tasks. We realize that the results are comparative, but the detailed result discussion is done in the upcoming Section~\ref{sec:discussion}.

\begin{table}
\caption{Accuracy score of cross-modal forced matching task comparing Learnable PINS~\cite{nagrani2018learnable}, SVHF-Net~\cite{nagrani2018seeing} and SSNet.}
\label{tab:nway}
\centering
\begin{tabular}{|cccc|}
\hline
Inputs & Learnable PINS.    & SVHF-Net  & Proposed SSNet \\
\hline
 & \multicolumn{3}{c|}{Voice $\rightarrow$ Face (\%)}\\
\hline
2 & 84 & 78 & 78\\
4 & 54 & 46 & 56\\
6 & 42 & 39 & 42\\
8 & 36 & 34 & 36\\
10 & 30 & 28 & 30\\
\hline
\end{tabular}
\end{table}


\begin{table*}
\caption{Cross-modal retrieval, $R@10$, results of single stream network trained under the framework employing both modalities as probe.}
\centering
\begin{tabular}{|ccccc|}
\hline
 & Random & Gender & Random & Gender  \\
 \hline\hline
&  Voice $\rightarrow$ Face ($R@10$) & & Face $\rightarrow$ Voice ($R@10$) & \\ 
 \hline
 & \multicolumn{4}{c|}{Seen-Heard}\\
\hline
Proposed SSNet & 36.27   & 37.20 & 50.00  & 51.20  \\
\hline
 & \multicolumn{4}{c|}{Unseen-Unheard}                            \\
 \hline
Proposed SSNet & 8.70  & -  & 13.20  & -  \\
\hline
\end{tabular}
\label{tab:results-retrieval}
\end{table*}

\begin{table}
\caption{Cross-modal Verification results on Seen-Heard and unseen-unheard configurations to illustrate the effect of proposed loss function.}
\begin{center}
\label{tab:ablation-study-uu}
\begin{tabular}{|ccc|}
\hline
Configuration & AUC \% & EER \% \\
 \hline
 & \multicolumn{2}{c|}{Seen-Heard}\\
 \hline
 $\lambda = 0.0$  & 81.2  & 26.3   \\
 $\lambda = 1.0$   & \textbf{91.1}  & \textbf{17.2}    \\
 \hline
 & \multicolumn{2}{c|}{Un-seen-Un-heard}\\
  \hline
$\lambda = 0.0$  & 72.6  & 33.6                       \\
$\lambda = 1.0$  &\textbf{78.8}  & \textbf{29.5}       \\
\hline
\end{tabular}
\end{center}
\end{table}

\begin{table}
\caption{Accuracy score of cross-modal forced matching task to illustrate the effect of proposed loss function.}
\label{tab:nway-ablation}
\centering
\begin{tabular}{|ccc|}
\hline
Inputs &  $\lambda = 0.0$     &  $\lambda = 1.0$   \\
\hline
 & \multicolumn{2}{c|}{Voice $\rightarrow$ Face (\%)}\\
\hline
2 & 73 & 78\\
4 & 49 & 56 \\
6 & 38 & 42 \\
8 & 34 & 36 \\
10 & 29 & 30 \\
\hline
\end{tabular}
\end{table}

\subsection{Cross-modal Retrieval}
In this section we evaluate the results of \textit{cross-modal retrieval} task employing both face and voice as probe with other modality at the retrieval end. We report results in terms of $R@K$ metric which evaluates the top $K$ retrieved results. Table~\ref{tab:results-retrieval} demonstrates quantitative results of our approach on the said task. We also perform retrieval conditioned on Gender as well for both formulations. Note that in Table~\ref{tab:results-retrieval} \textit{Random} corresponds to retrieval results on complete test set regardless of gender, nationality or age considerations. As an added experiment, we perform $R@10$ for \textit{gender}.

\subsection{Qualitative Evaluation}
\begin{figure*}[!t]
 \center
 \includegraphics[width=1.0\textwidth]{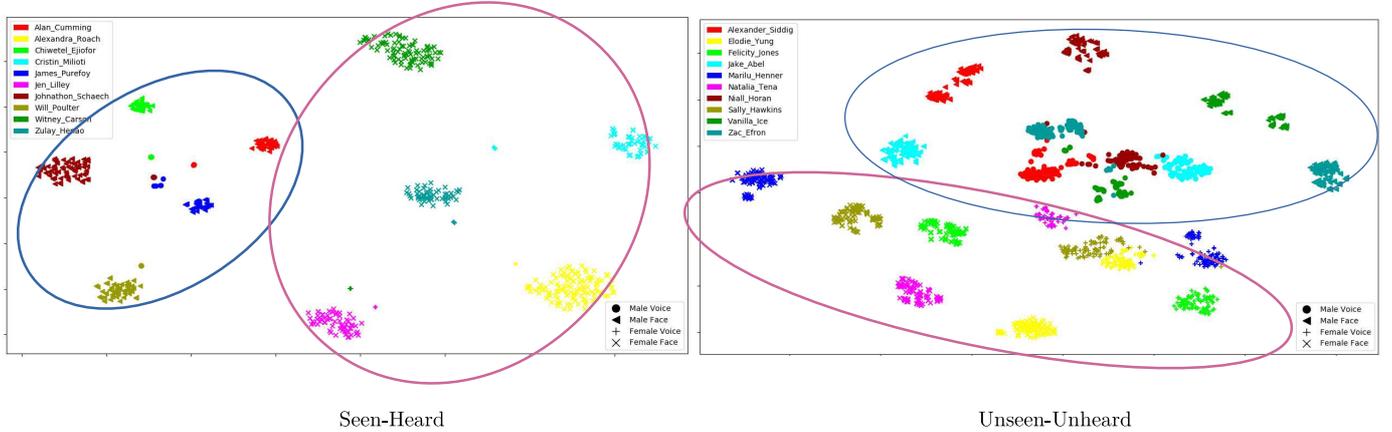}
  \caption{Visualization of learned voice and face embedding extracted from test set of the VoxCeleb dataset of $10$ identities. The pink oval encloses female entities while the male entities are enclosed in blue one. (Best viewed in color)  }
  \label{fig:tsne}
\end{figure*}

Fig.~\ref{fig:tsne} is tSNE~\cite{van2014accelerating} embedding result of learned features extracted from test set of VoxCeleb dataset of $10$ identities. We visualize learned embedding for both formulations i.e. \textit{seen-heard} and \textit{unseen-unheard}. Visual illustrations support the hypothesis that no pair selection knowledge at pre/post processing stage is required for network to learn mapping of identities in shared latent space. Given the faces and voices at input, the network learns to map both modalities conditioned at class information due to formulated loss function in Eq.~\ref{eq:centerloss-softmax}. Note that, \textit{unseen-unheard} formulation has very high level of difficulty even though results shown in Fig.~\ref{fig:tsne} are impressive.

\section{Result Discussion}
\label{sec:discussion}
In this section we discuss the results obtained for all three cross-modal tasks i.e. verification, matching and retrieval and performance of our approach. We compare our approach with~\cite{nagrani2018seeing} and~\cite{nagrani2018learnable} to list out the benefits of employing deep latent space learning framework as a training procedure along with the network coupled with formulated loss function. 
\subsection{No Pair Selection}
One of the main benefits on learning features in a shared latent space is no overhead of pair or triplet selection. As dataset increases exponentially over time, so does the overhead of pair or triplet selection. The proposed framework ensures that class information is leveraged to penalize distance between learned embedding.
\subsection{Ease of Fine tuning}
A biometrics system is supposed to be robust to age progression and capable of learning end-to-end from dynamic scenes as well. It is certain that fine tuning pre-existing systems can solve the problem of age progression. However, a major drawback of existing approaches is need of new pair wise information as soon as the input data changes. For every identity $i$, new pairs have to be selected to ensure robust representations. However, this is not the case with SSNet and the proposed loss function since it relies only on class information. Therefore, even though faces change over time, their parent class i.e. identity $i$ remains the same. Consequently, fine tuning just requires new input data with no pre/post processing and pretraining. 
\subsection{Component Modularity}
We argue that our approach is modular since single stream network coupled with loss functions are just components which are trained under the proposed framework. Therefore, it is inexpensive to switch between networks depending on the data.

\section{Ablation Study}
\label{sec:abl-study}
We experiment with the hyperparameter $\lambda$ which is used to couple conventional softmax with the proposed loss signal, in Eq~\ref{eq:centerloss-softmax}. When the value of $\lambda$ is set to $0$, the loss function is a special case where only softmax's penalization is utilized. Increasing value of $\lambda$ introduces the increasing effect of coupled penalization. We experiment with two values of $\lambda$ where we set it to $0$ and $1$ for two evaluation protocol of cross-modal verification and forced matching. The quantitative scores are reported in Table~\ref{tab:ablation-study-uu} and~\ref{tab:nway-ablation}. In these experiments, only $\lambda$ is varied otherwise joint formulation configuration is same as for previous experiments. These experiments explain the crucial need for penalization beyond softmax in the proposed setting and establish the effectiveness of penalization based on centers for tasks such as verification, matching and retrieval.

\section{Conclusion}
\label{sec:conclusion}
In this work, we presented a novel training procedure coupled with a single stream network capable of jointly embedding visual and audio signals into a shared latent space without any pairwise or triplet supervision. 
Furthermore, we introduced a novel supervision signal coupled with the single stream network to aid the joint projection of embedding. 
We demonstrated results on various cross-modal tasks: \textit{verification}, \textit{matching} and \textit{retrieval} considering several demographic factors such as age, gender, and nationality. We achieved state-of-the-art results on several tasks and comparable results on others despite simple strategy.

\bibliographystyle{IEEEtran}
\bibliography{bib}

\end{document}